\documentclass[letterpaper, 10 pt, conference]{ieeeconf}  

\IEEEoverridecommandlockouts                              

\usepackage[english]{babel}
\usepackage{times} 
\usepackage{cite}
\usepackage{hyperref}
\usepackage[font=footnotesize]{subcaption}
\usepackage{calrsfs}
\usepackage[nolist]{acronym}

\DeclareMathAlphabet{\pazocal}{OMS}{zplm}{m}{n}

\usepackage{cellspace}
\newcolumntype{D}{>{\hfill}N{3}{2}<{\hfill}}

\usepackage{ragged2e,tabularx, makecell}

\newcounter{Footnotecounter}
\newcommand{\setFootNote}[1][]{%
	\setcounter{Footnotecounter}{#1}%
}

\usepackage{graphicx} 
\usepackage{epsfig} 
\usepackage{pgfplots}
\usepackage{caption}
\usepackage{subcaption}

\pgfplotsset{compat=1.15}
\graphicspath{{./figure/}}

\usepackage[export]{adjustbox}

\makeatletter
\let\MYcaption\@makecaption
\makeatother

\makeatletter
\let\@makecaption\MYcaption
\makeatother

\usepackage{mathptmx} 
\usepackage{amsmath} 
\usepackage{amssymb}  
\usepackage{nicefrac}
\usepackage{mathtools}
\usepackage{optidef}
\usepackage{leftidx}
\usepackage{bm} 

\def\x#1{\texttt{\expandafter\string\csname#1\endcsname}&\expandafter$\csname#1\endcsname$}

\makeatletter%
\AfterPreamble{%
	\usepackage{hyperref}%
	\let\oldhypertarget\hypertarget%
	\renewcommand{\hypertarget}[2]{%
		\oldhypertarget{#1}{#2}%
		\protected@write\@mainaux{}{%
			\string\expandafter\string\gdef%
			\string\csname\string\detokenize{#1}\string\endcsname{#2}%
		}%
	}%
	\newcommand{\myhyperlink}[1]{%
		\hyperlink{#1}{\csname #1\endcsname}%
	}%
}
\makeatother%

\DeclareMathOperator*{\minimize}{minimize}
\DeclareMathOperator*{\maximize}{maximize}

\newcounter{Definition}
\newcounter{Theorem}
\makeatletter
\providecommand{\bigsqcap}{%
	\mathop{%
		\mathpalette\@updown\bigsqcup
	}%
}
\newcommand*{\@updown}[2]{%
	\rotatebox[origin=c]{180}{$\m@th#1#2$}%
}
\makeatother

\usepackage{tikz}
\pgfdeclarelayer{foreground}
\pgfsetlayers{background, main, foreground}
\usetikzlibrary{quotes, angles, backgrounds, arrows, automata, shapes, positioning, calc, through, 
spy, decorations.pathreplacing, decorations.markings}

\tikzset{
	imglabel/.style={
		rectangle,
		inner sep=2pt,
		text=black,
		minimum height=1em,
		text centered,
		fill=white,
		fill opacity=1.0,
		text opacity=1,
		anchor=south west,
	},
}

\tikzset{
	state/.style={
		rectangle,
		draw=black, very thick,
		minimum height=1.0em,
		text centered,
	},
}

\tikzset{
	on each segment/.style={
		decorate,
		decoration={
			show path construction,
			moveto code={},
			lineto code={
				\path [#1]
				(\tikzinputsegmentfirst) -- (\tikzinputsegmentlast);
			},
			curveto code={
				\path [#1] (\tikzinputsegmentfirst)
				.. controls
				(\tikzinputsegmentsupporta) and (\tikzinputsegmentsupportb)
				..
				(\tikzinputsegmentlast);
			},
			closepath code={
				\path [#1]
				(\tikzinputsegmentfirst) -- (\tikzinputsegmentlast);
			},
		},
	},
	mid arrow/.style={postaction={decorate,decoration={
				markings,
				mark=at position .5 with {\arrow[#1]{stealth}}
	}}},
}

\usepackage{siunitx}
\usepackage{todonotes}

\usepackage{algorithm}
\usepackage[noend]{algpseudocode}

\makeatletter
\def\BState{\State\hskip-\ALG@thistlm}
\makeatother

\newcommand\copyrighttext{%
	\small \begin{center} \color{red} \textcopyright\,2021 IEEE. Personal use of this material is 
		permitted. Permission from IEEE must be obtained for all other uses, in any current or 
		future 
		media, including reprinting/republishing this material for advertising or promotional 
		purposes, 
		creating new collective works, for resale or redistribution to servers or lists, or reuse 
		of 
		any copyrighted component of this work in other works. \end{center}}
\newcommand\copyrightnotice{%
	\begin{tikzpicture}[remember picture,overlay]
	\node[anchor=south,yshift=25.6cm] at (current page.south) 
	{\color{red}\fbox{\parbox{\dimexpr\textwidth-\fboxsep-\fboxrule\relax}{\copyrighttext}}};
	\end{tikzpicture}%
}

\title{\copyrightnotice \LARGE \bf
	A Multi-Layer Software Architecture for Aerial Cognitive Multi-Robot Systems in Power Line 
	Inspection Tasks
}

\author{Giuseppe Silano$^1$, Jan Bednar$^1$, Tiago Nascimento$^{1,2}$, Jesus Capitan$^3$, Martin 
Saska$^1$, and Anibal Ollero$^3$
	\thanks{$^1$Giuseppe Silano, Jan Bednar, Tiago Nascimento, and Martin Saska are with the 
	Faculty of Electrical Engineering, Czech Technical University in Prague, Czech Republic (email: 
	{\tt\small \{silangiu, jan.bednar14, pereiti1, martin.saska\}@fel.cvut.cz}).}
	\thanks{$^2$Tiago Nascimento is with the Lab of Systems Engineering and Robotics (LASER), 
	Department of Computer Systems, Universidade Federal da Paraiba, Brazil (email: {\tt\small 
	tiagopn@ci.ufpb.br}).}
	\thanks{$^3$Jesus Capitan and Anibal Ollero are with the GRVC Robotics Laboratory, University 
	of Seville, Spain (email: {\tt\small \{jcapitan, aollero\}@us.es}).}
	\thanks{This work was partially funded by the European Union's Horizon 2020 research and 
	innovation programme AERIAL-CORE under grant agreement no. 871479, by CTU grant no. 
	SGS20/174/OHK3/3T/13, and by the Czech Science Foundation (GAČR), within research project no. 
	20-10280S.}
}

\begin{document}
	
	\maketitle
	\thispagestyle{empty}
	\pagestyle{empty}
	
	
	\begin{acronym}
		\acro{CNN}[CNN]{Convolutional Neural Network}
		\acro{FOV}[FoV]{Field of View}
		\acro{FSM}[FSM]{Failure recovery and Synchronization jobs Manager}
		\acro{ICP}[ICP]{Iterative Closest Point}
		\acro{IR}[IR]{InfraRed}
		\acro{GNSS}[GNSS]{Global Navigation Satellite System}
		\acro{GPS}[GPS]{Global Positioning System}
		\acro{MAV}[MAV]{Micro-scale Unmanned Aerial Vehicle}
		\acro{MBZIRC}[MBZIRC 2020]{Mohamed Bin Zayed International Robotics Challenge 2020}
		\acro{MBZIRC17}[MBZIRC 2017]{Mohamed Bin Zayed International Robotics Challenge 2017}
		\acro{MIDGARD}[MIDGARD]{MAV Identification Dataset Generated Automatically in Real-world 
		Deployment}
		\acro{MOCAP}[mo-cap]{Motion Capture}
		\acro{MPC}[MPC]{Model Predictive Control}
		\acro{MRS}[MRS]{Multi-Robot System}
		\acro{ML}[ML]{Machine Learning}
		\acro{NN}[NN]{Neural Network}
		\acro{ROS}[ROS]{Robot Operating System}
		\acro{ROW}[ROW]{Rolling On Wire}
		\acro{RTK}[RTK]{Real-time Kinematic}
		\acro{SIL}[SIL]{Software-in-the-loop}
		\acro{STL}[STL]{Signal Temporal Logic}
		\acro{UAV}[UAV]{Unmanned Aerial Vehicle}
		\acro{UGV}[UGV]{Unmanned Ground Vehicle}
		\acro{UV}[UV]{UltraViolet}
		\acro{UVDAR}[UVDAR]{UltraViolet Direction And Ranging}
		\acro{UT}[UT]{Unscented Transform}
	\end{acronym}
	
	
	
	\begin{abstract}
		
		This paper presents a multi-layer software architecture to perform cooperative missions 
		with a fleet of quad-rotors providing support in electrical power line inspection 
		operations. The proposed software framework guarantees the compliance with safety 
		requirements between drones and human workers while ensuring that the mission is carried 
		out successfully. Besides, cognitive capabilities are integrated in the multi-vehicle 
		system in order to reply to unforeseen events and external disturbances. The feasibility 
		and effectiveness of the proposed architecture are demonstrated by means of realistic 
		simulations. 
		
	\end{abstract}
	
	
	
	
	
	
	
	
	
	
	
	
	
	\section{Introduction}
	\label{sec:introduction}
	
	Over the last two decades, the global energy demand has increased rapidly due to demographic 
	and economic growth, especially in emerging market areas. This has created new challenges for 
	electricity supply companies, which are constantly looking for new solutions to minimize the 
	frequency of power outages. Power failures are particularly critical when the environment and 
	public safety are at risk, e.g., for hospitals, sewage treatment plants, and telecommunication 
	systems. Damaged transmission lines, usually due to high winds, storms, or inefficient 
	inspection campaigns~\cite{ParkJFR2019}, is one of the major causes of power outages.
	
\begin{figure}
		\begin{center}
			\adjincludegraphics[width=0.4\textwidth, trim={{0.0\width} {.15\height} {0.0\width} 
				{.10\height}}, clip]{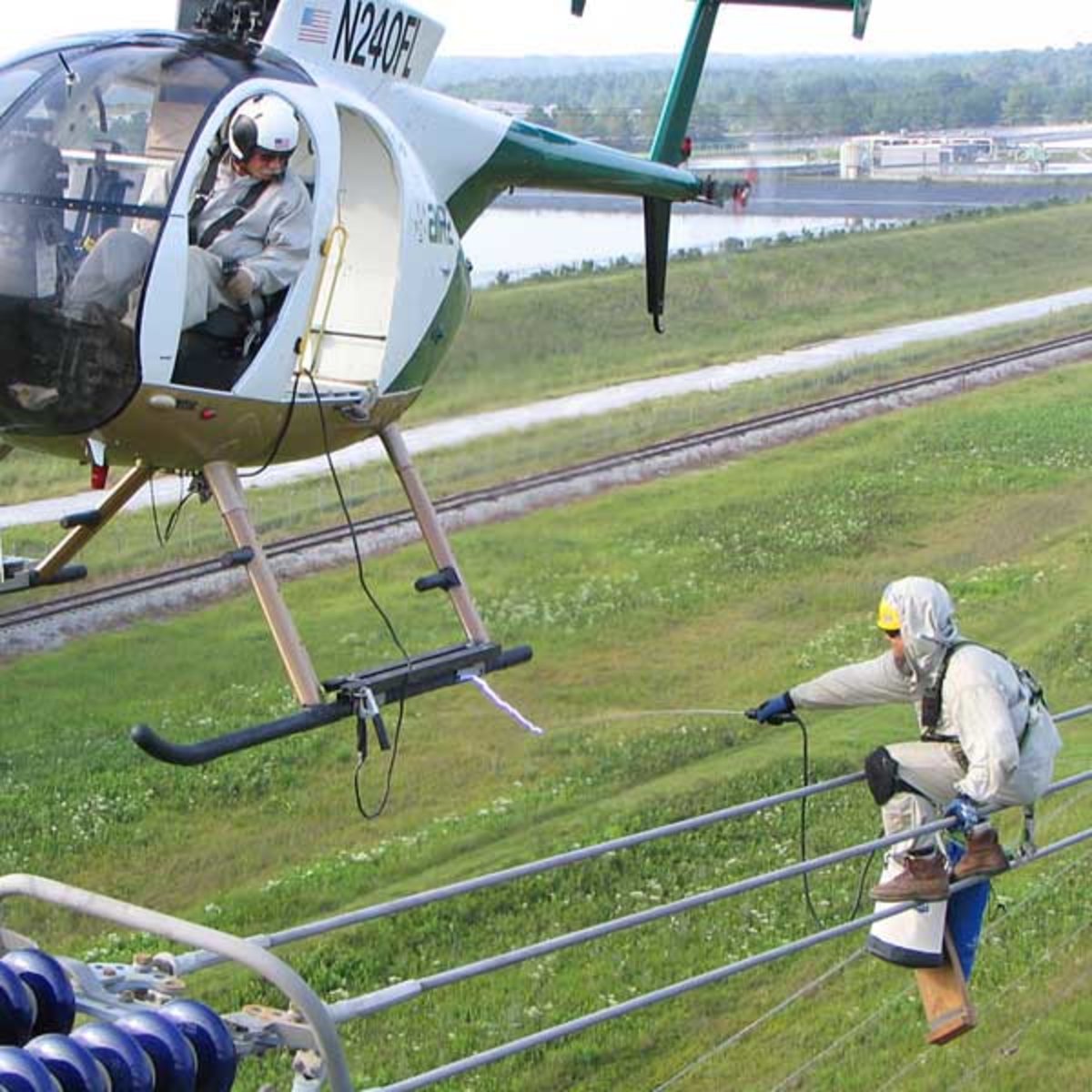}
		\end{center}
		\vspace*{-2mm}
		\caption{An operator climbing back to the helicopter after performing maintenance on the 
		conductors.}
		\label{fig:exampleInspectionMission}
	\end{figure}

	Nowadays, the most common strategy for reducing energy interruptions is to schedule periodic 
	maintenance activities by carrying out repairs and replacements on active lines (see, 
	Fig.~\ref{fig:exampleInspectionMission}). This is the most suitable method when system 
	integrity, reliability, and operating revenues are essential, and when the removal of a circuit 
	is not acceptable~\cite{ParkJFR2019}. Manned helicopters and experienced crews take care of 
	acquiring data over thousands of kilometers. Conductive suits, climbing harnesses, and arc 
	control rods prevent operators from getting shocked while working on transmission lines. 
	However, there are two major drawbacks of this approach: first, inspection and maintenance are 
	dangerous for operators who work close to power towers and operate on electrified lines; 
	second, these operations are extremely time-consuming and expensive (\$1,500 for a one-hour 
	flight) and prone to human error~\cite{Baik2018JIRS}.
	
	Therefore, there is a clear need for safe and practical techniques that enable more efficient 
	maintenance and inspection procedures in electrical power lines, in order to reduce potential 
	risks and costs for the distribution companies. Multiple solutions have been proposed in the 
	literature for automating this task~\cite{Martinez2018EAAI}, but the most promising and 
	flexible alternative is to use~\acp{UAV}, as they are capable of supporting inspection at 
	different levels~\cite{ParkJFR2019}. For instance, \acp{UAV} can inspect places of difficult 
	access, but they can also  monitoring human operations for safety purposes.
	
	However, the use of~\acp{UAV} for inspection tasks in power lines is particularly challenging 
	due to issues like their limited battery capacity, the strong electromagnetic interference 
	produced by power lines, and the presence of potential obstacles along the lines (e.g., 
	branches, vegetation, marker balls)~\cite{Baik2018JIRS}. Enhanced systems with cognitive 
	capabilities, e.g., based on novel perception sensors, such as event 
	cameras~\cite{GallegoTPAMI2020}, advanced data fusion techniques~\cite{WalterRAL2019}, or fast 
	on-line planning~\cite{Penicka2019RAL}, are of interest to address those complexities and to 
	accomplish the assigned mission safely and successfully. More precisely, it is key for the 
	system to integrate~\acp{UAV} capable of processing the acquired knowledge of the surrounding 
	environment and of planning and executing appropriate actions in reaction to unforeseen events 
	and external disturbances. Furthermore, these actions should be performed in a cooperative 
	manner by the fleet of \acp{UAV} while ensuring the compliance with safety requirements. 
	
	\begin{table*}
		\centering
		\def\arraystretch{0.8}
		\setlength\tabcolsep{4.0pt} 
		\begin{tabular}{c||c|c|c|c|c|c|c|c|c}
			\noalign{\smallskip}\hline
			\small platform & \begin{tabular}{@{}c@{}} \small open\\[-0.1em] \small 
			source\end{tabular} & \small modular & \small simulation & \begin{tabular}{@{}c@{}} 
			\small outside\\[-0.1em] \small laboratory\end{tabular} & \begin{tabular}{@{}c@{}} 
			\small cognitive\\[-0.1em] \small architecture\end{tabular} & \begin{tabular}{@{}c@{}} 
			\small multi-frame\\[-0.1em] \small localization\end{tabular}& \begin{tabular}{@{}c@{}} 
			\small rate\\[-0.1em] \small output\end{tabular} & \begin{tabular}{@{}c@{}} \small 
			software last\\[-0.1em] \small update\end{tabular} & \small reference\\
			\noalign{\smallskip}\hline\hline\noalign{\smallskip}
			\small \small MRS UAV system & \small + & \small + & \small + & \small + & \small + & 
			\small + & \small + & \small 2020 & \small \cite{Baca2020mrs}\\
			\small Aerostack & \small + & \small + & \small + & \small + & \small - & \small - & 
			\small - & \small 2020 & \small \cite{sanchez2016aerostack}\\
			\small XTDrone & \small + & \small + & \small + & \small - & \small - & \small - & 
			\small - & \small 2020 & \small \cite{xiao2020xtdrone}\\
			\small RotorS & \small + & \small + & \small +  & \small - & \small - & \small - & 
			\small + & \small 2020 & \small \cite{furrer2016rotors}\\
			\small ReCOPTER & \small + & \small - & \small - & \small - & \small - & \small -  & 
			\small - & \small 2015 & \small \cite{Abeywardena2015design}\\
			\small MAVwork & \small + & \small + & \small - & \small - & \small -  & \small - & 
			\small - & \small 2013 & \small \cite{Mellado2013mavwork}\\
			\noalign{\smallskip}\hline
		\end{tabular}
		\caption{Comparison of available open-source frameworks for~\ac{UAV}.}
		\label{tab:uav_systems}
	\end{table*}
	
	Versatile and reliable software architectures are essential to integrate these cognitive 
	multi-\ac{UAV} systems with many interconnected heterogeneous components (e.g., path planners, 
	control and computer vision algorithms, task managers). Therefore, in this paper, we propose 
	such an architecture to deal with multi-\ac{UAV} missions in the context of power line 
	inspection and human safety.
	
	
	
	\subsection{Related works}
	\label{sec:raltedWorks}
	
	\ac{UAV}~frameworks have been proposed over the last $15$ years as valuable tools for 
	inspection and surveillance purposes~\cite{Ollero2018RAM}, soil and field analysis and crop 
	monitoring~\cite{sanchez2016aerostack}. However, the design of a complete software stack for 
	fully autonomous multi-\ac{UAV} systems is still an open problem involving multiple 
	interconnected aspects, such as the design of guidance and navigation, control 
	systems~\cite{Jacquet2021RAL, Baca2020mrs}, and the development of  a reliable communication 
	network~\cite{Ryll2015TCST}. Several commercial and open-source projects have been proposed 
	over the years to develop ready-to-use hardware and software architectures for multi-rotor 
	vehicles~\cite{Cervera2019RAL}. Nevertheless, most of the shared code is not well documented, 
	making it difficult to reuse components, to add new features, or to replicate the research 
	results. These software projects do not offer guarantees on the reliability of the code or 
	customization possibilities.
	
	
	Table~\ref{tab:uav_systems} summarizes the most popular software frameworks available in the 
	literature. MAVwork~\cite{Mellado2013mavwork} was one of the first frameworks for aerial 
	vehicles in $2011$. Two years later, ReCOPTER~\cite{Abeywardena2015design, Sanchez-Lopez2016} 
	was released, which is an open-source framework for research and educational activities with 
	multi-rotor vehicles. Although the software was released as supporting material for the paper, 
	no further updates have been released since then, with the consequent issue for code 
	reusability~\cite{Cervera2019RAL}.
	The first up-to-date framework to be mentioned is the RotorS~\cite{furrer2016rotors} simulator. 
	This work provides Gazebo-based simulations for numerous heterogeneous vehicles (e.g., the 
	Ascending Firefly Hexarotor, the Parrot AR.Drone, the VoliroX). However, the control pipeline 
	features are too basic, with little potential to be applied to real-world conditions. 
	Another~\ac{UAV} framework is XTDrone~\cite{xiao2020xtdrone}, which offers a simulation testbed 
	with many complex functionalities, including simulation of onboard sensors and complex 
	localization systems. The control pipeline relies entirely on the PX4 embedded control 
	software, which limits its application to other hardware platforms. In contrast, the Aerostack 
	system~\cite{sanchez2016aerostack}, designed for the deployment of multi-rotor~\acp{UAV}, is 
	continuously being updated, and offers a straightforward transition from simulation to real 
	experiments. However, this system is based on the DJI flight controller, whose control inputs 
	are limited to orientation and thrust commands.  Furthermore, the system lacks the feature of 
	switching between multiple frames of reference. 
	
	
	
	
	\subsection{Contributions}
	\label{sec:contributions}
	
	In this paper, we propose a multi-layer software architecture designed for the AERIAL-CORE 
	European project\footnote{\url{https://aerial-core.eu}}, in which cognitive aerial platforms 
	are being developed inspired by the application of autonomous power line inspection. The 
	proposed software framework relies on the~\ac{UAV} platform in~\cite{Baca2020mrs} and is built 
	on top of  the~\ac{ROS} shell. The Nimbro 
	network\footnote{\url{https://github.com/ctu-mrs/nimbro_network}} is used to support 
	communication between vehicles by implementing the \textit{TCP Fast-Open protocol} and reducing 
	bandwidth using the \textit{libbz2} data compression algorithm. 
	
	In comparison to the frameworks described in Section~\ref{sec:raltedWorks}, the advantages are 
	threefold: (i) besides the RotorS and Aerostack systems, no other existing platform provides a 
	full-stack framework for multi-rotor~\acp{UAV} that is actively maintained and supports 
	simulation with a fleet of aerial vehicles; (ii) the integration of the Nimbro network ensures 
	that the~\ac{ROS} messages are transmitted at the published rate, a crucial requirement in 
	multi-\ac{UAV} applications; and (iii) none of the above-mentioned frameworks present cognitive 
	functionalities at the level required by the applications in the AERIAL-CORE project. For 
	instance, these cognitive capabilities should enable the system to adapt to information learned 
	online, e.g.,~\acp{UAV} failing or battery levels dropping faster than expected. 
	
	
	Gazebo simulations show the validity and effectiveness of the proposed platform, which is 
	released as 
	open-source\footnote{\setFootNote[\thefootnote]\url{https://github.com/ctu-mrs/icuas_2021_sw_architecture_acws}}.
	 These preliminary simulations showcase the capabilities and potential of the architecture. 
	Moreover, they demonstrate an advanced level of integration of the whole multi-\ac{UAV} system 
	in the AERIAL-CORE project.
	
	
	
	\section{Problem Description}
	\label{sec:problemDescription}
	
	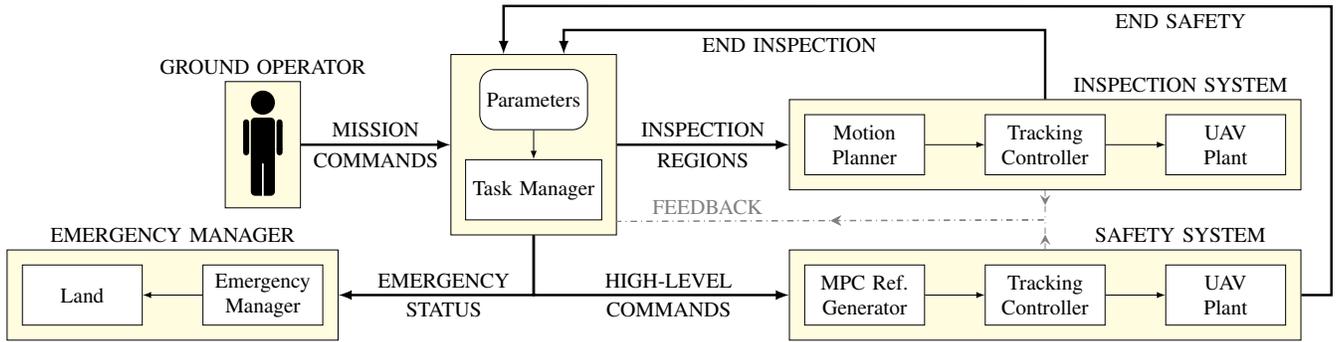
\begin{figure*}
		\begin{center}
			\scalebox{0.8}{
				\begin{tikzpicture}
				
				\node (groundOperator) at (0,-0.7) [draw, fill=yellow!15, rectangle, minimum 
				height=2.1cm, minimum width=1.25cm, text centered]{};
				\node at (0,0.6) [text centered]{GROUND OPERATOR};
				
				\node[circle,fill,minimum size=4mm] (head) {};
				\node[rounded corners=2pt,minimum height=1.3cm,minimum width=0.4cm,fill,below = 1pt 
				of head] (body) {};
				\draw[line width=1mm,round cap-round cap] ([shift={(2pt,-1pt)}]body.north east) 
				--++(-90:6mm);
				\draw[line width=1mm,round cap-round cap] ([shift={(-2pt,-1pt)}]body.north 
				west)--++(-90:6mm);
				\draw[thick,white,-round cap] (body.south) --++(90:5.5mm);
				
				\node (boxTaskManager) at (4.5,-0.7) [rectangle, draw, fill=yellow!15, minimum 
				width=2.75cm, minimum height=3cm]{}; 
				\node (taskManagerParameters) at (4.5,0.05) [draw, fill=white, rounded 
				corners=0.25cm, minimum height=1cm, text centered]{Parameters};
				\node (taskManager) at (4.5,-1.45) [draw, rectangle, fill=white, minimum 
				height=1cm, minimum width=1.5cm, text centered]{Task Manager};
				
				\draw[-latex] (taskManagerParameters) -- (taskManager);
				
				\node (boxInspectionManager) at (13,-0.7) [rectangle, draw, fill=yellow!15, minimum 
				width=8.5cm, minimum height=1.5cm]{}; 
				\node (pathPlannerInspection) at (10,-0.7) [draw, rectangle, fill=white, minimum 
				height=1cm, minimum width=1cm, text centered, text width=5em]{Motion\\Planner};
				\node (trackingController) at (13,-0.7) [draw, rectangle, fill=white, minimum 
				height=1cm, minimum width=1cm, text centered, text width=5em]{Tracking\\Controller};
				\node (UAVPlant) at (16,-0.7) [draw, rectangle, fill=white, minimum height=1cm, 
				minimum width=1cm, text centered, text width=5em]{UAV\\Plant};
				
				\draw[-latex] (pathPlannerInspection) -- (trackingController);
				\draw[-latex] (trackingController) -- (UAVPlant);
				
				\node at (15.25,0.3) [text centered]{INSPECTION SYSTEM};
				
				\node (boxSafetyManager) at (13,-3.2) [rectangle, draw, fill=yellow!15, minimum 
				width=8.5cm, minimum height=1.5cm]{}; 
				\node (MPCReferenceGenerator) at (10,-3.2) [draw, rectangle, fill=white, minimum 
				height=1cm, minimum width=1cm, text centered, text width=5em]{MPC Ref.\\Generator};
				\node (trackingControllerMPC) at (13,-3.2) [draw, rectangle, fill=white, minimum 
				height=1cm, minimum width=1cm, text centered, text width=5em]{Tracking\\Controller};
				\node (UAVPlantMPC) at (16,-3.2) [draw, rectangle, fill=white, minimum height=1cm, 
				minimum width=1cm, text centered, text width=5em]{UAV\\Plant};
				
				\draw[-latex] (MPCReferenceGenerator) -- (trackingControllerMPC);
				\draw[-latex] (trackingControllerMPC) -- (UAVPlantMPC);
				
				\node at (15.25,-2.2) [text centered]{SAFETY SYSTEM};
				
				\node (boxEmergencyManager) at (-1.5,-3.2) [rectangle, draw, fill=yellow!15, 
				minimum width=5.5cm, minimum height=1.5cm]{}; 
				\node (EmergencyManager) at (0,-3.2) [draw, rectangle, fill=white, minimum 
				height=1cm, minimum width=1cm, text centered, text width=5em]{Emergency\\Manager};
				\node (Land) at (-3,-3.2) [draw, rectangle, fill=white, minimum height=1cm, minimum 
				width=1cm, text centered, text width=5em]{Land};
				
				\draw[-latex] (EmergencyManager) -- (Land);
				
				\node at (-1.5,-2.2) [text centered]{EMERGENCY MANAGER};
				
				\draw[-latex, line width=1.15pt] (groundOperator) -- node[above]{MISSION} 
				node[below]{COMMANDS} (boxTaskManager);
				\draw[-latex, line width=1.15pt] (boxTaskManager) -- node[above]{INSPECTION} 
				node[below]{REGIONS} (boxInspectionManager);
				\draw[-latex, line width=1.15pt] (boxTaskManager) |- (boxSafetyManager);
				\node at ($ (boxSafetyManager.west) - (2,0) $) [above]{HIGH-LEVEL};
				\node at ($ (boxSafetyManager.west) - (2,0) $) [below]{COMMANDS};
				\draw[-latex, line width=1.15pt] (boxTaskManager) |- (boxEmergencyManager);
				\node at ($ (boxEmergencyManager.east) + (1.75,0) $) [above]{EMERGENCY};
				\node at ($ (boxEmergencyManager.east) + (1.75,0) $) [below]{STATUS};
				
				\draw[-latex, line width=1.15pt] (boxInspectionManager.north) -- ($ 
				(boxInspectionManager.north) + (0,0.5) $) |- ( $(boxTaskManager.north) + (0.5,0.4) 
				$) -- ($ (boxTaskManager.north) + (0.5,0) $);
				\node at ($ (boxInspectionManager.north) + (-4.25,1.15) $) [below]{END INSPECTION};
				
				\draw[-latex, line width=1.15pt] (boxSafetyManager.east) -- ($ 
				(boxSafetyManager.east) + (0.5,0) $) -- ($ (boxSafetyManager.east) + (0.5,4.8) $) 
				-- ( $(boxTaskManager.north) - (0.5,-0.8) $) -- ($ (boxTaskManager.north) - (0.5,0) 
				$);
				\node at ($ (boxInspectionManager.north) - (-2.25,-1.05) $) [above]{END SAFETY};
				
				\draw[dashdotted, gray, postaction={on each segment={mid arrow=gray}, line 
				width=1.15pt}] (boxInspectionManager.south) -- ($ (boxInspectionManager.south) + 
				(0,-0.5)$) -- ( $ (boxTaskManager) + (1.375,-1.275) $);
				\draw[dashdotted, gray, postaction={on each segment={mid arrow=gray}, line 
				width=1.15pt}] (boxSafetyManager.north) -- ($ (boxSafetyManager.north) + (0,0.5)$) 
				-- ( $ (boxTaskManager) + (1.375,-1.275) $);
				\node at ($ (boxSafetyManager.north) + (-6.65,0.7)$) [right, gray]{FEEDBACK};
				
				\end{tikzpicture}
			}
		\end{center}
		\caption{Proposed software platform architecture. Arrows represent the data exchanged among 
			blocks.}
		\label{fig:softwareArchitecture}
	\end{figure*}
	
\begin{figure}
		\centering
		
		\begin{subfigure}{0.45\columnwidth}
			\centering
			\adjincludegraphics[width=0.97\textwidth, trim={{0.37\width} {0.3\height} {0.35\width} 
			{0.3\height}}, clip]{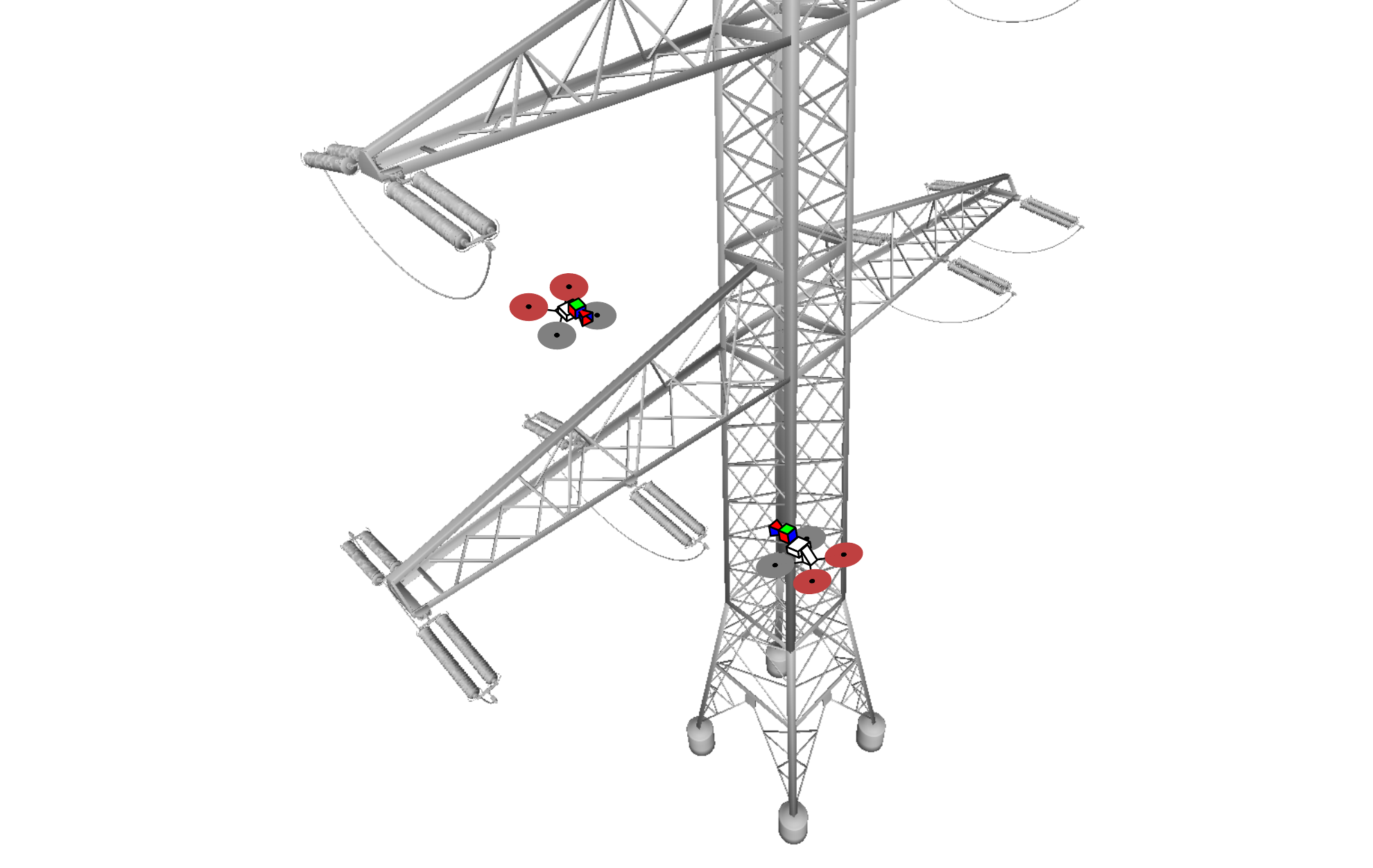}
			\caption{}
			\label{fig:inspectionACW}
		\end{subfigure}
		\hspace{0.5mm}
		%
		\begin{subfigure}{0.45\columnwidth}
			\centering
			\adjincludegraphics[width=1.05\textwidth, trim={{0.0\width} {0.1\height} {0.0\width} 
			{0.0\height}}, clip]{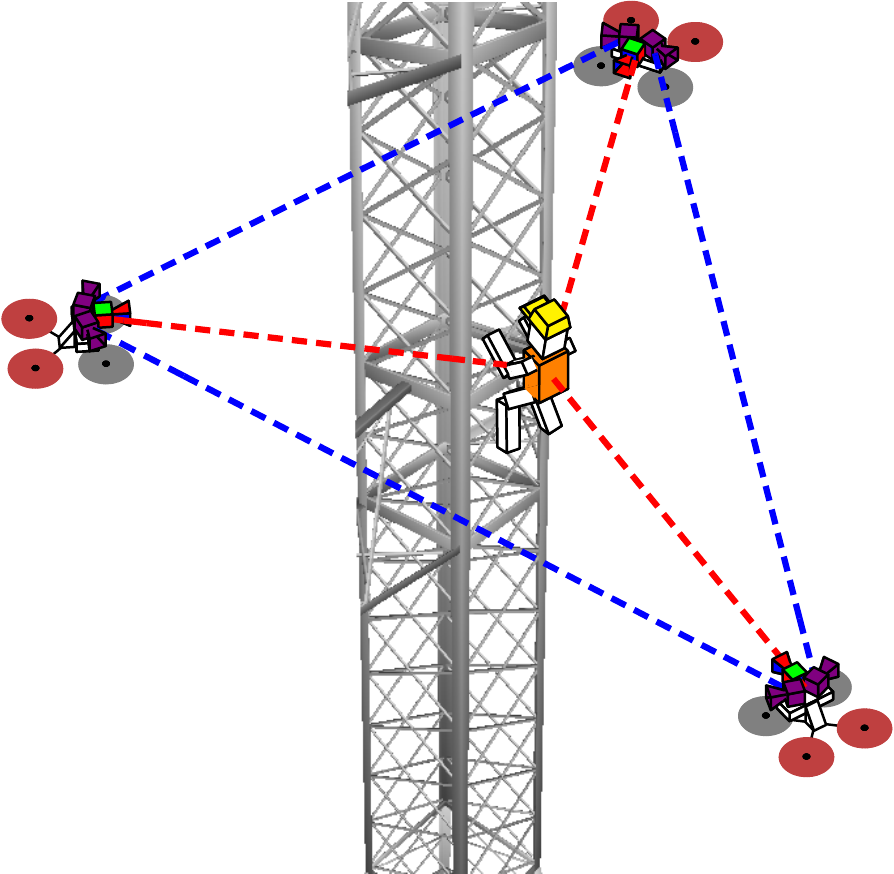}
			\caption{}
			\label{fig:safetyACW}
		\end{subfigure}
		\caption{From left to right: \textit{inspection} and \textit{safety} scenarios. Onboard 
		cameras and sensors acquire data of the power equipment. A mutual localization system helps 
		to maintain the formation avoiding contact with the tower.}
	\end{figure}

	
	Two tasks of interest are considered: (i) \textit{inspection}, where a fleet of 
	multi-rotor~\acp{UAV} carries out a detailed investigation of power equipment autonomously, 
	helping the human workers to acquire views of the power tower that are not easily accessible 
	(see, Fig.~\ref{fig:inspectionACW}); and (ii) \textit{safety}, where a formation of~\acp{UAV} 
	provides the supervising team with a view of the humans working on the power tower in order to 
	monitor their status and to ensure their safety (see, Fig.~\ref{fig:safetyACW}).
	
	In both tasks, visual sensors are essential to perform the required work. In the~\ac{UAV} 
	configuration, cameras are mounted in an \textit{eye-in-hand} configuration, i.e., rigidly 
	attached to the body frame. For the \textit{inspection} task, cameras capture images of the 
	power tower searching for damage to the mechanical structure and for failures of the electrical 
	components. In the \textit{safety} task, a visual servoing scheme is implemented to keep track 
	of the movements and actions of the human workers throughout the entire operation. Camera 
	images are also used to mutually localize the~\acp{UAV} in the surrounding 
	environment~\cite{WalterRAL2019}. 
	
	Finally, we assume that the~\acp{UAV} operate in an environment represented by a previously 
	acquired map, including the position of the power towers and other potential obstacles. We also 
	assume that the~\acp{UAV} are equipped with the necessary sensors and software for precise 
	self-location and state estimation~\cite{Baca2020mrs}.
	
	
	
	\section{Software Framework}
	\label{sec:softwareFramework}
	
	The software platform is organized into three layers of abstraction that provide high 
	modularity and flexibility, while reflecting the problem description: the \textit{task manager} 
	(Section~\ref{sec:taskManager}), and the \textit{inspection} 
	(Section~\ref{sec:inspectionMission}) and the \textit{safety} (Section~\ref{sec:safetyMission}) 
	systems. The framework controls the behavior of the~\acp{UAV} and can be visualized as a chain 
	of various software components working together to fulfill the mission specifications. The 
	\textit{task manager} and the \textit{motion planner} block inside the \textit{inspection} 
	system are placed at a ground station. In contrast, all other blocks and layers of the proposed 
	system run onboard the~\acp{UAV} as a distributed system. The framework enables the~\acp{UAV} 
	through the \textit{task manager system} to execute the \textit{inspection} and \textit{safety} 
	tasks. Continuous feedback from the~\acp{UAV} guarantees a certain degree of reliability 
	against unexpected behaviors stopping the mission in case of problems. 
	Figure~\ref{fig:softwareArchitecture} presents the proposed software architecture. 

	
	
	\subsection{Task manager}
	\label{sec:taskManager}
	
	The \textit{task manager} is the core cognitive block of the system and it is in charge of 
	implementing high-level behaviors, by computing multi-\ac{UAV} cooperative plans and allocating 
	tasks to different \acp{UAV} coping with their heterogeneity and their battery levels. 
	Each~\ac{UAV} is assumed to have different onboard sensors or configurations in order to be 
	more suitable for a specific task. This block implements planning algorithms on demand, so that 
	the team of~\acp{UAV} is able to support the human operators working on an electrical tower; 
	but it is also capable of re-planning online to react to new learnt circumstances. 
	
	Therefore, the \textit{task manager} has the cognitive capability to learn policies online 
	using available information and to react to unexpected events not present in the initial plan 
	(e.g.,~\ac{UAV} failures, shorter flight time than expected, etc.). This is achieved through a 
	high-level planner that computes initial plans given~\ac{UAV} constraints in terms of battery 
	limits and heterogeneous capabilities. Then, the block monitors the plan execution 
	continuously, integrating information perceived by the~\acp{UAV}, in order to re-plan online if 
	the initial plan is no longer feasible. Besides, \emph{emergency} maneuvers are commanded for 
	those~\acp{UAV} failing, so that they land safely.  
	
	The ground operator is the first input of the system, providing a complete mission to be 
	performed by the fleet of~\acp{UAV}. Given that mission, the \textit{task manager} computes a 
	plan to implement the required actions with the team. These tasks consist of parametric 
	high-level commands for different \textit{inspection} and \textit{safety} activities. As basic 
	primitives, the~\acp{UAV} can be commanded to stay idle or to go to a known recharging station. 
	Also, in case of a low battery level, an emergency landing is commanded. The other two 
	high-level commands are: (i) to perform an \textit{inspection} task, with the regions to be 
	inspected encoded as a sequence of target regions; and (ii) to perform a \textit{safety} task, 
	with the identifier of the worker to be monitored and the geometry of the formation (i.e., the 
	distance to the worker, the viewing angles, and the inter-\ac{UAV} angles) as parameters. These 
	high-level commands are sent out to the lower layers of the architecture, which are then in 
	charge of implementing the corresponding behaviors by means of low-level controllers that deal 
	with multi-\ac{UAV} navigation and formation control. 
	
	The \textit{task manager} also receives feedback information from the~\acp{UAV} about their 
	battery level, localization and target state estimation, in order to generate its learnt 
	representation of the environment. This information is used to decide when a~\ac{UAV} has 
	failed and needs to be landed urgently, and when the current plan is not feasible anymore and 
	new task assignments are required, for instance to replace a~\ac{UAV} that may run out of 
	battery before finishing its current task. 
	
	Regarding the underlying algorithm for high-level planning, our implementation consists of a 
	Behavior Tree that encodes the constraints imposed by~\ac{UAV} battery levels and heterogeneous 
	capabilities, trying to allocate tasks to~\acp{UAV} with the objective of minimizing total 
	travel time. 
	
	
	
	\subsection{Inspection system}
	\label{sec:inspectionMission}
	
	The \textit{inspection} system is in charge of computing feasible and constrained trajectories 
	for a fleet of $q$ multi-rotors that have been assigned an \textit{inspection} task together. 
	This is done by leveraging on~\ac{STL}~\cite{Silano2021ICRARAL} specifications to perform the 
	inspection of a power tower. Such a logic allows for planning and executing appropriate 
	actions, starting from (possibly vague) high-level task specifications (e.g., the~\acp{UAV} 
	should reach the goal within $10$ time units while always avoiding obstacles). In 
	particular,~\ac{STL} can be used to describe planning objectives that are more complex than 
	point-to-point planning algorithms (e.g., A$^\star$, RRT$^\star$).  
	
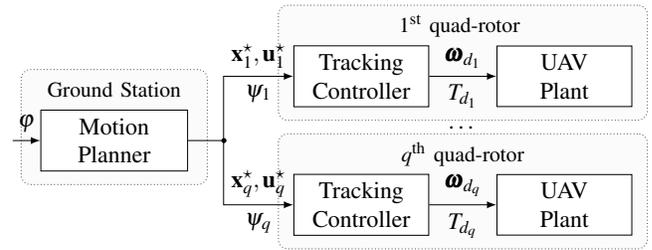
\begin{figure}
		\begin{center}
			\scalebox{0.9}{
				\begin{tikzpicture}
				
				\node (MultiRobots-Box) at (0,0.2) [fill=gray!3,rounded corners, draw=black!70, 
				densely 
				dotted, minimum height=1.7cm, minimum width=2.75cm]{}; 
				
				\node (MotionPlanner) at (0,0) [text centered, fill=white, draw, rectangle, minimum 
				width=1.5cm, text width=5.5em]{Motion\\Planner};
				
				\draw[-latex] ($(MotionPlanner) - (1.5,0)$) -- node[above]{$\varphi$} 
				(MotionPlanner);
				
				\draw[fill=black] ($ (MotionPlanner.east) + (0.5,0) $) arc(-180:180:0.03);
				\node (GroundStation) at (0,0.75) [text centered]{\small Ground Station};
				
				\node (Drone-Box1) at (5.15,1.15) [fill=gray!3,rounded corners, draw=black!70, 
				densely 
				dotted, minimum height=1.7cm, minimum width=5.45cm]{}; 
				\node (TrackingController1) at (3.65,0.95) [fill=white, draw, rectangle, text 
				centered, 
				text width=5em]{Tracking\\Controller};
				\node (UAVPlant1) at (6.65,0.95) [fill=white, draw, rectangle, text centered, text 
				width=5em]{UAV\\Plant};
				\node (Drone-Box1-Text) at (5.1,1.7) [text centered]{\small $1$\textsuperscript{st} 
					quad-rotor};	
				
				\draw[-latex] (TrackingController1) -- node[above]{$\bm{\omega}_{d_1}$} 
				node[below]{$T_{d_1}$} 
				(UAVPlant1);
				\draw[-latex] (MotionPlanner.east) -- ($ (MotionPlanner.east) + (0.525,0) $) 
				-- ($ (MotionPlanner.east) + (0.525,0.95) $) -- node[above]{$\mathbf{x}^\star_1, 
					\mathbf{u}^\star_1$} node[below]{$\psi_1$} (TrackingController1.west);
				
				%
				
				\node (Drone-BoxN) at (5.15,-0.75) [fill=gray!3,rounded corners, draw=black!70, 
				densely 
				dotted, minimum height=1.7cm, minimum width=5.45cm]{};
				\node (Dots2) at (5.15,0.2) [text centered]{\dots};
				\node (TrackingControllerN) at (3.65,-0.95) [fill=white, draw, rectangle, text 
				centered, 
				text width=5em]{Tracking\\Controller};
				\node (UAVPlantN) at (6.65,-0.95) [fill=white, draw, rectangle, text centered, text 
				width=5em]{UAV\\Plant}; 	
				\node (Drone-BoxN-Text) at (5.15,-0.2) [text centered]{\small 
				$q$\textsuperscript{th} 
					quad-rotor};
				
				\draw[-latex] (TrackingControllerN) -- node[above]{$\bm{\omega}_{d_q}$} 
				node[below]{$T_{d_q}$} 
				(UAVPlantN);
				\draw[-latex] ($ (MotionPlanner.east) + (0.525,0) $) -- ($ 	
				(MotionPlanner.east) + (0.525,-0.95) $) -- node[above]{$\mathbf{x}^\star_q, 
					\mathbf{u}^\star_q$} node[below]{$\psi_q$} (TrackingControllerN.west);
				
				\end{tikzpicture}
			}
			\caption{\textit{Inspection} control scheme. Starting from~\ac{STL} formula $\varphi$, 
			the motion planner generates the trajectories $\left( \mathbf{x}_i^\star, 
			\mathbf{u}_i^\star \right)$ and the heading angles $\psi_i$, with $i=\{1, \dots, q\}$, 
			for the $q$~\acp{UAV}. A trajectory tracking controller supplies the desired angular 
			velocities $\bm{\omega}_{d_i}$ and thrust $T_{d_i}$ commands for the~\acp{UAV}.}
			\label{fig:overallSystem}
		\end{center}
		\vspace{-1em}
	\end{figure}

	An optimization problem~\eqref{eq:optimizationProblemInspectionACW} is formulated to generate 
	feasible dynamic trajectories that satisfy these specifications and also take vehicle 
	constraints into account, i.e., the maximum velocity and acceleration of the vehicles, by using 
	the motion primitives defined in~\cite{Mueller2015TRO}. The planner~\cite{Silano2021ICRARAL} 
	enables the formulation of complex missions that avoid obstacles and maintain a safe distance 
	between~\acp{UAV} while performing the inspection within a given time. When the~\acp{UAV} reach 
	the target regions, they start collecting images and videos and acquiring data from the onboard 
	sensors. Both the inspection time and target regions as well as other high-level commands, such 
	as stay idle or recharge, are defined by the \textit{task manager} and encoded in the~\ac{STL} 
	formula ($\varphi$) before the mission starts. Figure~\ref{fig:overallSystem} describes the 
	overall \textit{inspection} control architecture.
	
	Let $T_s \in \mathbb{R}_{\geq 0}$ and $T \in \mathbb{R}_{\geq 0}$ be the sampling period and 
	the trajectory duration, respectively; we can write the time interval as the vector 
	$\mathbf{t}=(0, T_s, \dots, N T_s)^\top \in \mathbb{R}^{N+1}$, where $NT_s = T$ and 
	$\mathbf{t}_k$, $k \in \mathbb{N}_{\geq 0}$, denote the $k$-element of the vector $\mathbf{t}$. 
	Similarly, let us define the state $\mathbf{x}$ and control $\mathbf{u}$ sequences of the 
	system as $\mathbf{x}_k = (\mathbf{p}_k^{(1)}, \mathbf{v}_k^{(1)}, \mathbf{p}_k^{(2)}, 
	\mathbf{v}_k^{(2)}, \mathbf{p}_k^{(3)}, \mathbf{v}_k^{(3)})^\top$ and $\mathbf{u}_k = 
	(\mathbf{a}_k^{(1)}, \mathbf{a}_k^{(2)}, \mathbf{a}_k^{(3)})^\top$, where $\mathbf{p}_k^{(j)}$, 
	$\mathbf{v}_k^{(j)}$, $\mathbf{a}_k^{(j)}$, with $j=\{1,2,3\}$, represent the vehicle's 
	position, velocity, and acceleration at time instant $k$ along the $j$-axis for the inertial 
	frame. Finally, let us consider the~\ac{STL} formula $\varphi$ and its smoothed robustness 
	version $\tilde{\mathbf{\rho}}_\varphi(\mathbf{x}, \mathbf{t}_k)$. The optimization problem can 
	be formalized as follows:
	\begin{equation}\label{eq:optimizationProblemInspectionACW}
	\begin{split}
	&\maximize_{\mathbf{p}^{(j)}, \mathbf{v}^{(j)},\,\mathbf{a}^{(j)}}\;
	{\tilde{\rho}_\varphi (\mathbf{p}^{(j)}, \mathbf{v}^{(j)} )} \\
	&\quad \,\;\, \text{s.t.}~\quad\; \lvert \mathbf{v}^{(j)}_k \rvert \leq 
	\mathbf{v}^{(j)}_\mathrm{max}, \lvert \mathbf{a}^{(j)}_k \vert  \leq 
	\mathbf{a}^{(j)}_\mathrm{max}, 
	\\
	&\,\;\;\;\, \qquad \quad~[18,\text{eq.}~(2)], \forall k=\{0,1, \dots, N-1\}
	\end{split},
	\end{equation}
	where $\mathbf{v}^{(j)}_\mathrm{max}$ and $\mathbf{a}^{(j)}_\mathrm{max}$ are the desired 
	maximum values of velocity and acceleration along the motion, respectively. The heading angles 
	($\psi$) are provided as a constant reference for each target region. Further details are 
	available in~\cite{Silano2021ICRARAL, Baca2020mrs}.

	\subsection{Safety system}
	\label{sec:safetyMission}
	
	The \textit{safety} system aims to control a team of $g$ \acp{UAV} that have been assigned a 
	\textit{safety} task to provide views of a human working on the power tower in order to monitor 
	his status while ensuring compliance with safety requirements. A~\ac{MPC} framework based 
	on~\cite{Saska2020AR, Kratky2020RAL, Jacquet2021RAL} deals with generating an optimal 
	trajectory ($\mathbf{p}^\star$, $\mathbf{v}^\star$, and $\mathbf{a}^\star$) for the vehicles 
	avoiding collisions with the power tower and obstacles, considering the~\acp{UAV} physical 
	constraints (i.e., maximum velocity and acceleration), and keeping a viewing angle with respect 
	to the worker in the camera. 
	
	A real-time implementation of the optimization problem~\eqref{eq:optimizationProblem} is 
	considered to cope with a \textit{safety} mission. Through the \textit{task manager}, the 
	ground operator can provide high-level commands to the aerial vehicles so that they change 
	their viewing angles or the shape of their formation (i.e., their distance to the worker or 
	their inter-\ac{UAV} angles). The mutual localization system in~\cite{WalterRAL2019} helps to 
	maintain the \ac{UAV} formation by complementing the GPS vehicle positions when electromagnetic 
	interference generated by the power tower are not negligible.
	
	
	\begin{figure}
		\begin{center}
			\scalebox{0.65}{
				\begin{tikzpicture}
				
				\node (Region1) at (6,0.1) [fill=gray!3, rounded corners, draw=black!70, densely 
				dotted, minimum height=1.7cm, minimum width=2.25cm]{}; 
				\node (Region2) at (9.5,-0.5) [fill=gray!3, rounded corners, draw=black!70, densely 
				dotted, minimum height=3.0cm, minimum width=2.25cm]{}; 
				\node (UAVPlant-Text) at (9.5,0.75) [text centered]{UAV Plant};
				\node (RateController-Text) at (6,0.725) [text centered]{Autopilot};
				
				\node (MPCReferenceGenerator) at (-1,0) [draw, rectangle, minimum width=1cm, 
				minimum height=1cm, text centered, text width=7em, fill=white]{MPC 
				Reference\\Generator};
				\node (ReferenceController) at (2.5,0) [draw, rectangle, minimum width=1cm, minimum 
				height=1cm, text centered, text width=5em, fill=white]{Reference\\Controller};
				\node (RateController) at (6,0) [draw, rectangle, minimum width=1cm, minimum 
				height=1cm, text centered, text width=5em, fill=white]{Rate\\Controller};
				\node (Actuators) at (9.5,0) [draw, rectangle, minimum width=1cm, minimum 
				height=1cm, text centered, text width=5em, fill=white]{Actuators};
				\node (Sensors) at (9.5,-1.25) [draw, rectangle, minimum width=1cm, minimum 
				height=1cm, text centered, text width=5em, fill=white]{Sensors};
				\node (Localization) at (9.5,-2.75) [draw, rectangle, minimum width=1cm, minimum 
				height=1cm, text centered, text width=5em, fill=white]{Localization};
				\node (StateEstimator) at (6,-2.75) [draw, rectangle, minimum width=1cm, minimum 
				height=1cm, text centered, text width=5em, fill=white]{State\\Estimator};
				
				\draw[-latex] (MPCReferenceGenerator) -- node[above]{$\mathbf{x}^\star$, 
				$\mathbf{u}^\star$} node[below]{$\psi$} (ReferenceController);
				\draw[-latex] (ReferenceController) -- node[above]{$\bm{\omega}_d$, $T_d$} 
				node[below]{\SI{100}{\hertz}} (RateController);
				\draw[-latex] (RateController) -- node[above]{$\bm{\tau}_d$} node[below]{$\approx$ 
				\SI{1}{\kilo\hertz}} (Actuators);
				\draw[-latex] (Sensors.south) -- (Localization.north);
				\draw[-latex] (Localization.west) -- (StateEstimator.east);
				\draw[latex-] (StateEstimator.north) -- node[left]{$\mathbf{R}$, $\bm{\omega}$} 
				(RateController.south);
				\draw[-latex] (StateEstimator.west) -- ($ (StateEstimator.west) - (2.5,0) $) -- 
				node[text width=7em]{$\zeta$, $\mathbf{R}$, $\bm{\omega}$\\ \;\SI{100}{\hertz}} 
				(ReferenceController.south);
				\draw[-latex] ($ (MPCReferenceGenerator.south) - (0,1)$) node[below, text centered, 
				text width=5em]{High-level\\Commands} -- (MPCReferenceGenerator.south);
				
				\end{tikzpicture}
			}
		\end{center}
		\caption{\textit{Safety} control architecture. The \textit{MPC Reference Generator} 
		supplies the trajectory $\bigl( \mathbf{x}^\star$, $\mathbf{u}^\star \bigr)$ and the 
		heading angle $\psi$ to the \textit{Reference Controller}, which outputs the thrust $T_d$ 
		and angular velocities $\bm{\omega}_d$ for the autopilot that computes the propellers speed 
		$\bm{\tau}_d$ for the \textit{Actuators}. A \textit{State Estimator} provides the~\ac{UAV} 
		translation and rotation ($\zeta$, $\mathbf{R}$, $\bm{\omega}$).}
		\label{fig:controlSchemeSafetyACW}
	\end{figure}
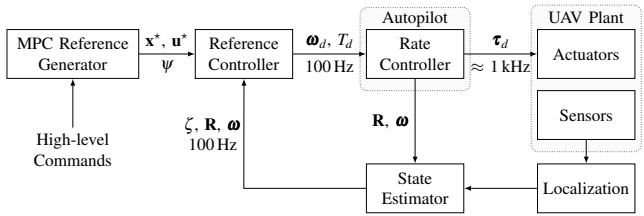
	
	Let us consider a continuous-time dynamical system $\pazocal{H}$ and its discrete time version 
	$x_{k+1} = f(x_k, u_k)$, where $x_k$, $x_{k+1} \in X \subset \mathbb{R}^n$ are the current 
	state and the next state of the system, respectively, $u \in U \subset \mathbb{R}^m$ is the 
	control input, and $f \colon X \times U \rightarrow X$ is differentiable in both the arguments. 
	The initial state is denoted by $x_0$ and takes values from some initial set $X_0 \subset 
	\mathbb{R}^n$. Let us also assume that $z_k$ is the state vector of the perception system 
	(e.g., the projections of the worker's 3D points on the image plane, the target position, 
	etc.), and $\sigma$ a set of parameters characterizing them. The perception and the vehicle 
	states are coupled through the~\ac{UAV} dynamics, namely $z_{k+1}= f_p(x_k, u_k, \sigma)$. 
	Hence, the discrete-time optimization problem over a receding horizon~$T_h$, sampled in $W$ 
	shooting points, at a given instant $t$, can be expressed as:
	\begin{equation}\label{eq:optimizationProblem}
	\begin{split}
	&\minimize_{\mathbf{x}, \mathbf{u}, \mathbf{z}} \;\;\; J \bigl(\mathbf{x}, \mathbf{u} \bigr) + 
	J_p \bigl( \mathbf{z} \bigr) \\
	&\quad\;\;\text{s.t.} \qquad  r(\mathbf{x}_k, \mathbf{u}_k, \mathbf{z}_k) = 0, \qquad   \\
	& \qquad \;\;\; \qquad h(\mathbf{x}_k, \mathbf{u}_k, \mathbf{z}_k) \leq 0 
	\end{split},
	\end{equation}
	where $J(\mathbf{x}, \mathbf{u})$ and $J_p(\mathbf{z})$ are the action and perception objective 
	functions, respectively, while $r(\mathbf{x}_k, \mathbf{u}_k, \mathbf{z}_k)$ and 
	$h(\mathbf{x}_k, \mathbf{u}_k, \mathbf{z}_k)$ represent equality and inequality constraints 
	that the solution should satisfy for perception, action, or both simultaneously, respectively. 
	Roughly speaking, we encode the action objective $J(\mathbf{x}, \mathbf{u})$ ensuring the 
	minimum distance between the \ac{UAV} and the human worker, while being compliant with safety 
	requirements (i.e., maintaining a certain distance between \acp{UAV} and worker, bounding 
	maximum velocity and acceleration of the vehicle). Such requirements are included as hard 
	constraints in the $r(\mathbf{x}_k, \mathbf{u}_k, \mathbf{z}_k)$ and $h(\mathbf{x}_k, 
	\mathbf{u}_k, \mathbf{z}_k)$ functions. First and second time derivative of tracking error are 
	also considered to avoid discontinuities in the \ac{UAV} behavior. On the other side, we 
	integrate the visibility constraint in $J_p(\mathbf{z})$ such that the visibility cone 
	originated by the camera view\footnote{The \textit{pinhole camera model} is taken into account 
	for the vision sensor.}, especially its horizontal and vertical projections, includes at the 
	best the human features. The optimization problem relies on the assumption that the system is 
	deferentially flat. This allows for simplification of the optimization problem, transforming 
	the nonlinear dynamics of the $g$~\acp{UAV} in an equivalent linear description of the system. 
	Figure~\ref{fig:controlSchemeSafetyACW} describes the overall system architecture, while more 
	details are available in~\cite{Saska2020AR, Kratky2020RAL}.
	
	
	
	\section{Simulation Results}
	\label{sec:numericalEvaluations}
	
\begin{figure}
		\centering
		\begin{tikzpicture}
		
		\node[anchor=south west,inner sep=0] (img) at (0,0) { 
			\adjincludegraphics[width=0.475\textwidth, trim={{0.0\width} {0.0\height} {0.0\width} 
			{0.08\height}}, clip]{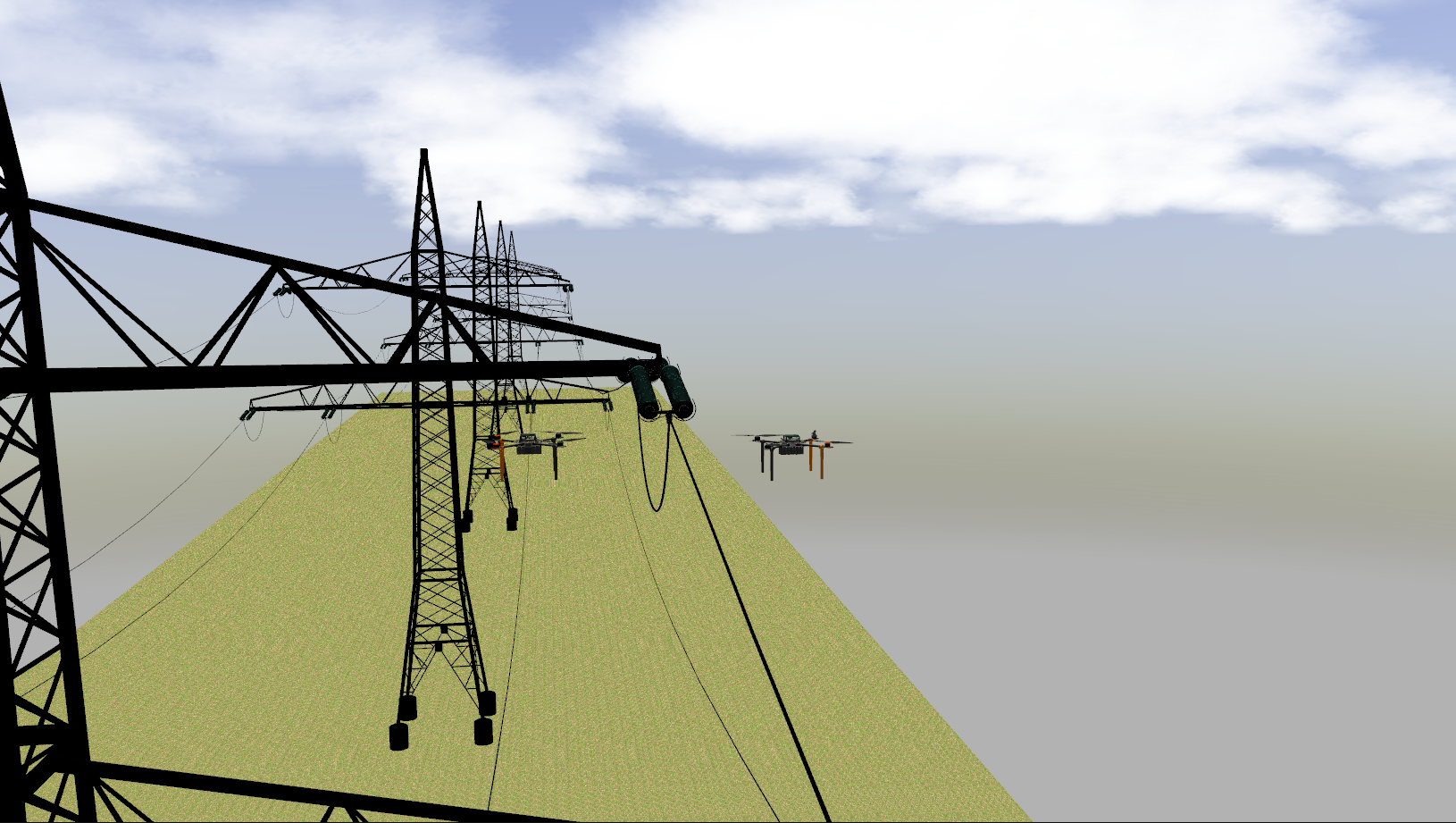}};
		\begin{scope}[x={(img.south east)},y={(img.north west)}]
		
		\draw [white, ultra thick] (0.3625, 0.485) circle (0.06);
		\draw [white, ultra thick] (0.545, 0.485) circle (0.06);
		
		\end{scope}
		\end{tikzpicture}
		\vspace*{-1mm}
		\caption{Snapshot of the \textit{inspection} scenario. Solid circles show the~\acp{UAV} 
		approaching an insulator.}
		\label{fig:inspectionScenarioSnapshot}
	\end{figure}

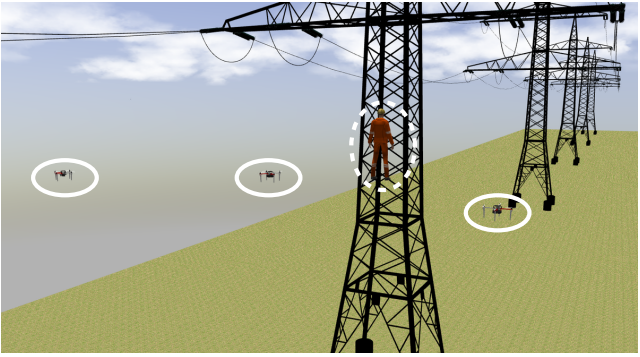
\begin{figure}
		\centering
		\begin{tikzpicture}
		
		\node[anchor=south west,inner sep=0] (img) at (0,0) { 
			\adjincludegraphics[width=0.475\textwidth, trim={{0.0\width} {0.08\height} {0.0\width} 
			{0.0\height}}, clip]{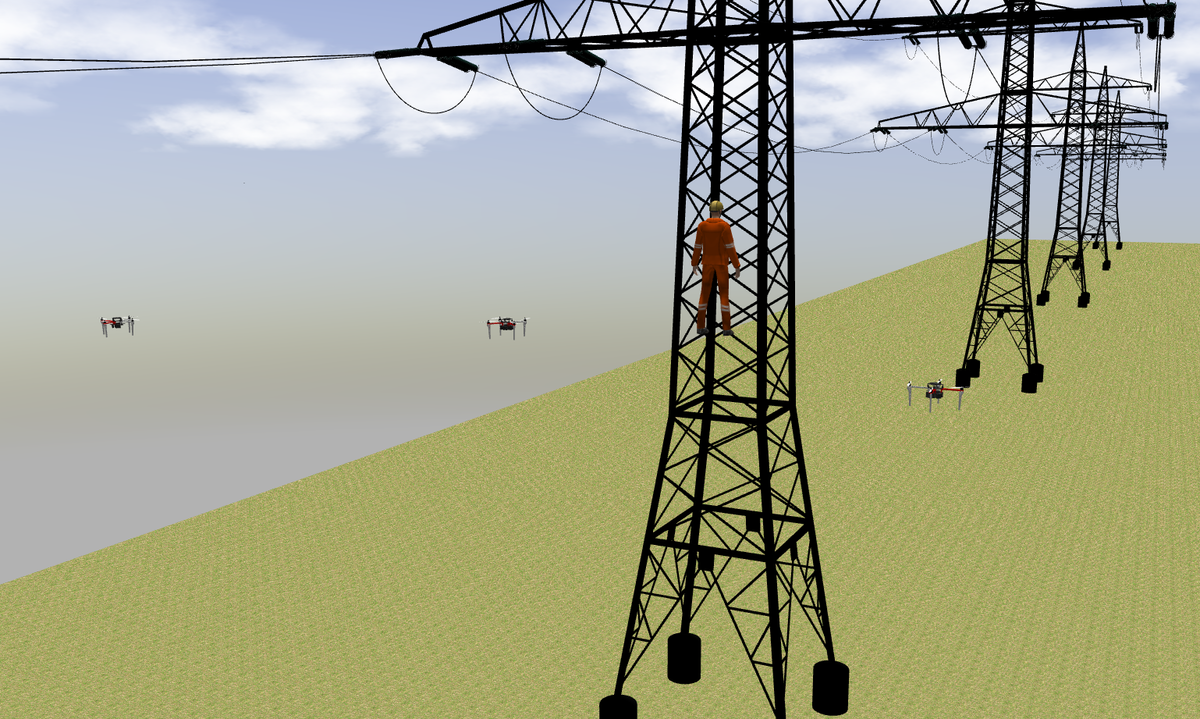}};
		\begin{scope}[x={(img.south east)},y={(img.north west)}]
		
		\draw [white, ultra thick] (0.1, 0.50) circle (0.05);
		\draw [white, ultra thick] (0.42, 0.50) circle (0.05);
		\draw [white, ultra thick] (0.78, 0.40) circle (0.05);
		
		\draw [white, dashed, ultra thick] (0.6, 0.59) ellipse (0.05 and 0.125);
		
		\end{scope}
		\end{tikzpicture}
		\vspace*{-1mm}
		\caption{Snapshot of the \textit{safety} scenario. Solid and dashed circles indicate the 
		\acp{UAV} and the operator, respectively.}
		\label{fig:safetyScenarioSnapshot}
	\end{figure}

	In this section, we show some preliminary simulation results to demonstrate the feasibility and 
	the effectiveness of the proposed software architecture. In particular, we simulated the system 
	in a realistic scenario by using the Gazebo robotic simulator, exploiting the advantages of 
	Software-in-the-loop simulations~\cite{Silano2019SMC}. Our objective is also to show an 
	advanced level of integration of the architecture. The framework was coded by using the Melodic 
	Moreina release of~\ac{ROS} with the optimization problems formulated using the CASADI 
	library\footnote{\url{https://web.casadi.org}} and NLP\footnote{\url{http://cvxr.com}} and 
	CVXGEN\footnote{\url{https://cvxgen.com}} as solvers. All simulations were performed on a 
	laptop with an i7-8565U processor (1.80 GHz) and 32GB of RAM running on Ubuntu $18.04$. 
	Figures~\ref{fig:inspectionScenarioSnapshot} and~\ref{fig:safetyScenarioSnapshot} depict 
	snapshots of an \textit{inspection} and a \textit{safety} mission, respectively. Videos with 
	the two simulations can be found at~\url{http://mrs.felk.cvut.cz/software-architecture-acws}. 
	
	Both the \textit{inspection} and \textit{safety} scenarios consist of a series of power towers, 
	each one with up to twelve insulators. The towers are $\SI{20}{\meter}$ high with a radius of 
	$\SI{15}{\meter}$. The presence of wires between the towers is also considered to simulate a 
	scenario quite close to the real application. The STL models and the mesh files have also been 
	made available\textsuperscript{\theFootnotecounter}. For the sake of simplicity and ease of 
	experimentation, we considered only one tower and six target regions for the 
	\textit{inspection} mission, but this does not imply a loss of generality of the architecture. 
	
	In the \textit{inspection} scenario (see, Fig.~\ref{fig:inspectionScenarioSnapshot}), the 
	\textit{task manager} provides the inspection order of the insulators considering the battery 
	limits and the heterogeneous capabilities of the vehicles before starting the mission. It also 
	sets the desired maximum values of velocity and acceleration of the vehicles. Then, the 
	optimization problem~\eqref{eq:optimizationProblemInspectionACW} is solved to provide the 
	dynamic feasible trajectories for the~\acp{UAV}. In the presented experimental results, a 
	safety distance of $\SI{1}{\meter}$ is maintained between the~\acp{UAV}, while the maximum 
	velocity $\mathbf{v}_\mathrm{max}^{(j)}$ and acceleration $\mathbf{a}_\mathrm{max}^{(j)}$ are 
	set to $\SI{3}{\meter\per\second}$ and $\SI{2.5}{\meter\per\square\second}$, respectively. The 
	maximum velocity and acceleration values were chosen in accordance with the camera constraints 
	to avoid producing blurred images. Besides, the safety distance guarantees some robustness with 
	respect to the GPS accuracy in order to avoid drift that may impact the system location and 
	lead the~\acp{UAV} to potential collisions.
	
	In the \textit{safety} scenario (see, Fig.~\ref{fig:safetyScenarioSnapshot}), three 
	multi-rotors fly in a formation while tracking a human working on a tower. The optimization 
	problem~\eqref{eq:optimizationProblem} running onboard the vehicles guarantees the compliance 
	with safety requirements (i.e., avoiding collisions with the power tower and with obstacles 
	along the path) while it ensures that the mission is carried out successfully. To evaluate and 
	demonstrate the applicability of the proposed software framework, we simulated high-level 
	commands from the ground operator that, through the \textit{task manager}, change the 
	viewpoints of the formation by moving the vehicles around the tower. Note that the \textit{task 
	manager} precludes the~\acp{UAV} from running out of battery by reacting on time, sending them 
	for emergency landing and reassigning their tasks to others. Further details on the onboard 
	navigation system and sensors are available in~\cite{Baca2020mrs}.

	
	
	\section{Conclusions}
	\label{sec:conclusion}
	
	This paper has presented a multi-layer software architecture for encoding and supporting 
	cooperative power line inspection operations with a fleet of~\acp{UAV}. Cognitive capabilities 
	have been considered for the safe and successful accomplishment of the assigned missions. The 
	architecture is designed around a set of software components that handle the current states of 
	the system, assign high-level tasks, and monitor the progress of the fully autonomous mission, 
	while ensuring compliance with safety requirements. Simulations in Gazebo have demonstrated the 
	feasibility and the effectiveness of the proposed framework, aiming towards the fulfillment of 
	real-world tests. Future work will include the integration of more challenging cognitive 
	capabilities, such as human interaction and gesture recognition to learn humans' intentions, 
	and lead to field experiments. Furthermore, we plan to investigate on planning algorithms that 
	can deal with uncertainties in task execution and workers' intentions.
	
	
	
	
	\bibliographystyle{IEEEtran}
	\bibliography{bib_short.bib}
	
\end{document}